\documentclass[letterpaper, 10 pt, conference]{ieeeconf}  

\IEEEoverridecommandlockouts                              

\usepackage{amsmath} 
\usepackage{amssymb}  
\usepackage{graphicx}

\usepackage{algorithm}
\usepackage{algpseudocode}
\usepackage{booktabs}
\usepackage{multirow}
\usepackage{subfig}

\usepackage{amsmath,amsfonts,bm}

\def\eqref#1{equation~\ref{#1}}

\def\1{\bm{1}}

\def\vtheta{{\bm{\theta}}}
\def\va{{\bm{a}}}

\def\vs{{\bm{s}}}

\def\vx{{\bm{x}}}
\def\vy{{\bm{y}}}

\def\mA{{\bm{A}}}
\def\mB{{\bm{B}}}

\DeclareMathAlphabet{\mathsfit}{\encodingdefault}{\sfdefault}{m}{sl}
\SetMathAlphabet{\mathsfit}{bold}{\encodingdefault}{\sfdefault}{bx}{n}

\newcommand{\E}{\mathbb{E}}

\newcommand{\R}{\mathbb{R}}

\DeclareMathOperator*{\argmin}{arg\,min}

\providecommand{\ie}{\emph{i.e.,} }
\providecommand{\eg}{\emph{e.g.,} }
\providecommand{\parab}[1]{\noindent\textbf{#1}}

\newcommand\diam{\mathrm{diam}}
\newtheorem{theorem}{Theorem}[section]

\newtheorem{definition}{Definition}
\newtheorem{assumption}{Assumption}

\usepackage{hyperref}
\title{\LARGE \bf
Safe Reinforcement Learning for Legged Locomotion 
}

\author{
Tsung-Yen Yang$^{1,2}$
Tingnan Zhang$^{2}$
Linda Luu$^{2}$
Sehoon Ha$^{2,3}$
Jie Tan$^{2}$
Wenhao Yu$^{2}$

\thanks{
$^{1}$TY is with Princeton University~{\tt\small ty3@princeton.edu}
$^{2}$TY, TZ, LL, SH, JT, and WY are with Google Research~{\tt\small\{jimmyyang, tingnan, luulinda, sehoonha, jietan, magicmelon\}@google.com} 
$^{3}$SH is with Georgia Institute of Technology~{\tt\small sehoonha@gatech.edu}
}
}

\begin{document}

\maketitle
\thispagestyle{empty}
\pagestyle{empty}

\begin{abstract}
Designing control policies for legged locomotion is complex due to the under-actuated and non-continuous robot dynamics. Model-free reinforcement learning provides promising tools to tackle this challenge. However, a major bottleneck of applying model-free reinforcement learning in real world is safety. In this paper, we propose a safe reinforcement learning framework that switches between a safe recovery policy that prevents the robot from entering unsafe states, and a learner policy that is optimized to complete the task. The safe recovery policy takes over the control when the learner policy violates safety constraints, and hands over the control back when there are no future safety violations. We design the safe recovery policy so that it ensures safety of legged locomotion while minimally intervening in the learning process. Furthermore, we theoretically analyze the proposed framework and provide an upper bound on the task performance. We verify the proposed framework in four locomotion tasks on a simulated and real quadrupedal robot: efficient gait, catwalk, two-leg balance, and pacing. On average, our method achieves 48.6\% fewer falls and comparable or better rewards than the baseline methods in simulation. When deployed it on real-world quadruped robot, our training pipeline enables 34\% improvement in energy efficiency for the efficient gait, 40.9\% narrower of the feet placement in the catwalk, and two times more jumping duration in the two-leg balance. Our method achieves less than five falls over the duration of 115 minutes of hardware time. \footnote{Video is included in the submission and the project website: \url{ https://sites.google.com/view/saferlleggedlocomotion/}}
\end{abstract}

\section{Introduction}

The promise of deep reinforcement learning (RL) in solving complex and high-dimensional problems autonomously has attracted much interest among robotics researchers. However, effectively training an RL policy requires exploring a large set of robot states and actions, including many that are not safe for the robot. This is especially true for systems that are inherently unstable such as legged robots. One way to leverage RL for robotics problems is to learn the policy in computer simulation and then deploy it in the real world \cite{lee2020learning, akkaya2019solving}. However, this requires addressing the challenging sim-to-real gap. Another approach to tackle this issue is to directly learn or fine-tune a control policy in the real-world \cite{ha2020learning, haarnoja2018learning}, with the main challenge being ensuring safety during learning. Our work falls into this category by introducing a safe RL framework for learning legged locomotion while satisfying the safety constraints during training.

We formulate the problem of safe locomotion learning in the context of safe RL. Inspired by prior work~\cite{thananjeyan2021recovery, yu2020protective, bharadhwaj2020conservative}, our learning framework adopts a two-policy structure: a \emph{safe recovery policy} that recovers robots from near-unsafe states, and a \emph{learner policy} that is optimized to perform the desired control task. Our safe learning framework switches between the safe recovery policy and the learner policy to prevent the learning agent from safety constraint violations (\eg falls).

Different from prior methods that learn a safety critic function which predicts the possibility of safe violations~\cite{thananjeyan2021recovery, yu2020protective, bharadhwaj2020conservative}, we propose a model-based approach to determine when to switch between the two policies based on the knowledge about the system dynamics.
In reality, it is often the case that the designer has some knowledge of the system dynamics at hand.
Our goal is to \textit{exploit this knowledge to design a safety mechanism without relying on a black-box approach} (\eg neural networks). 
More specifically, we first define a \textit{safety trigger set} that includes states where the robot is close to violating safety constraints but can still be saved by a safe recovery policy.
When the learner policy takes the robot to the safety trigger set, we switch to the safe recovery policy, which drives the robot back to safe states. We then determine when to switch back to the learner policy by leveraging an approximate dynamics model of the robot (\eg centroidal dynamics model for legged robots) to rollout the planned future robot trajectory: if the predicted future states are all in the safe states, we will hand the control back to learner policy, otherwise we will keep using the safe recovery policy (see Fig. \ref{fig:set} for illustration). 
Such switch criteria allow the learning agent to explore near safety-violation regions while minimally intervening in the learning process.

{
\begin{figure}[t]
\centering
{
\textbf{Catwalk} \\
\vspace{+1mm}
\includegraphics[width=0.95\linewidth]{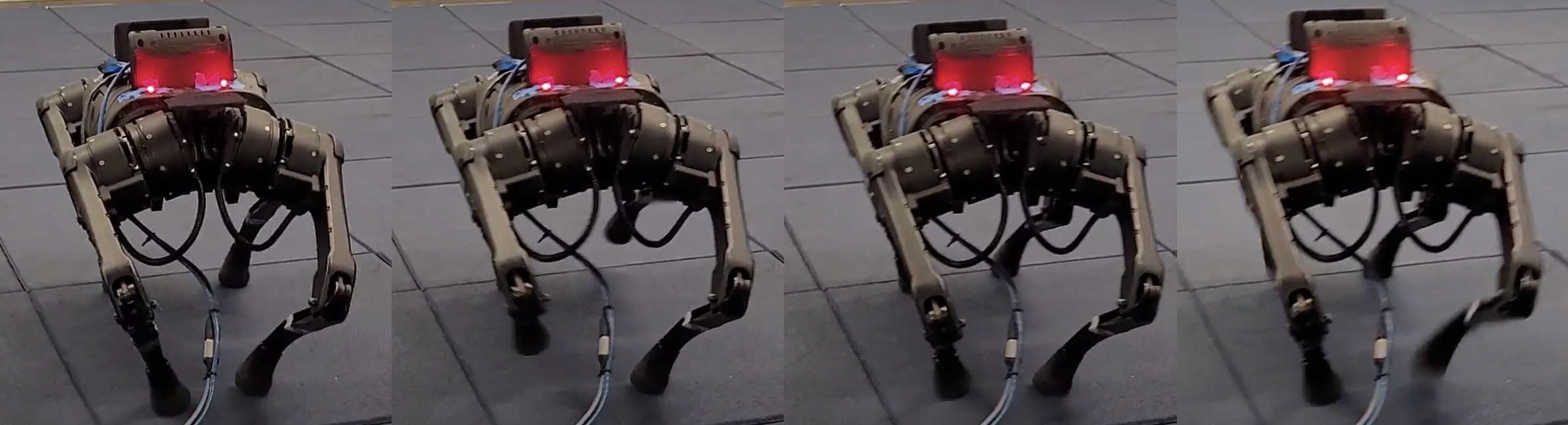} \\
\textbf{Two-leg balance}\\
\vspace{+1mm}
\hspace{+2mm}\includegraphics[width=0.95\linewidth]{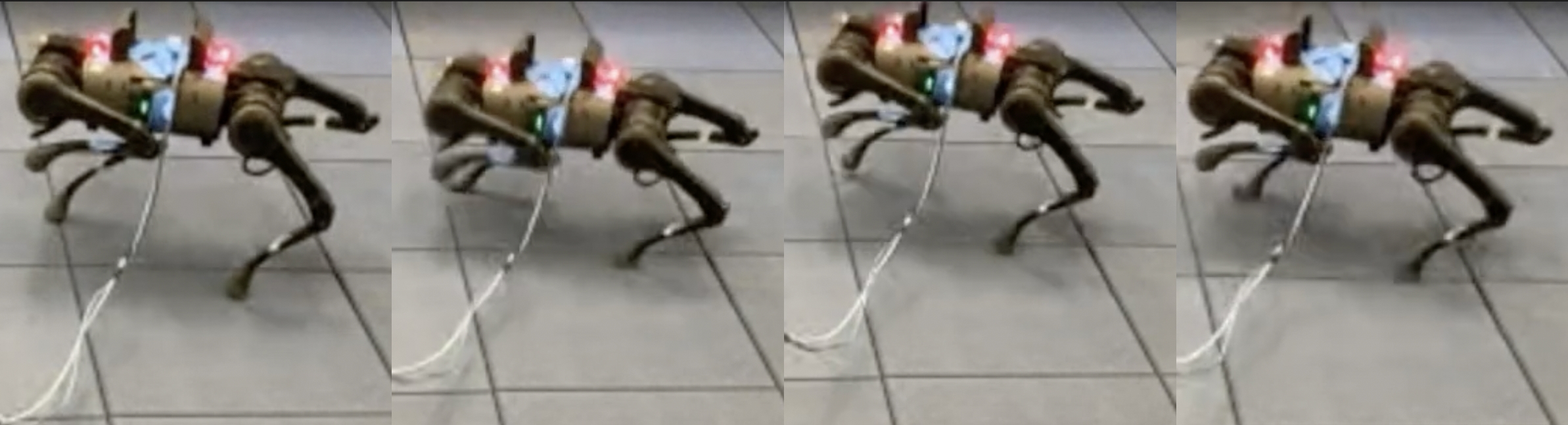}
}
\vspace{-2mm}
\caption{
We evaluate our algorithm in legged locomotion tasks: catwalk and two-leg balance.
The narrow feet placement and standing with two legs lead to falling easily during the learning process.
Our algorithm overcomes these challenges--outperforming prior methods in terms of safety violations and learning efficiency both in simulation and the real world.
}
\label{fig:algorithm}
\vspace{-7mm}
\end{figure}
}

The contributions are as follows.
\textbf{(1)} We provide a theoretical analysis of the proposed framework and derive performance bounds under system dynamics error since perfect knowledge about the system dynamics is hard to obtain.
Specifically, 
%
we construct a regret-type of bound that compares the learning outcome of the agent to the one with full knowledge about the system dynamics.
This helps us to quantify the effect of dynamics error on learning performance. 
\textbf{(2)} To evaluate the performance of the framework, we compare the proposed safe learning framework with state-of-the-art safe RL algorithms on four simulated and real legged locomotion tasks: efficient gait, catwalk, two-leg balance, and pacing (see Fig. \ref{fig:algorithm}).
In all cases, our algorithm achieves comparable or superior performance to prior approaches, averaging 48.6\% fewer falls in simulation and zero or near-zero falls in the real world.
In addition, we show that this framework can effectively reduce the number of uses of the safe recovery policy over prior approaches \cite{thananjeyan2021recovery}, which improves learning speed.

\section{Related Work}

\parab{Safe RL.}
Learning constraint-satisfying policies has been explored in the context of safe RL~\cite{garcia2015comprehensive,hasanbeig2020cautious,junges2016safety,jansen_et_al:LIPIcs:2020:12815,chow2018lyapunov,bharadhwaj2020conservative,srinivasan2020learning}.
In model-free RL, in which we do not know the system dynamics, prior work either designs policy update approaches~\cite{achiam2017constrained, yang2020projection}, adds a weighted copy of the cost function in the reward objective~\cite{tessler2018reward,chow2019lyapunov,fujimoto2019benchmarking,stooke2020responsive}, modifies output actions~\cite{dalal2018safe,avni2019run},
or predict the probability of being unsafe~\cite{fu2018risk,zheng2020constrained}.
Despite requiring little prior knowledge about the robotics system, these methods suffer from safety constraint violations during the learning process, especially during the initial learning stage where random exploration of the neural network policy is performed.
In contrast, our approach exploits the knowledge of the system dynamics to do safe planning.
Many prior works in model-based RL use a Gaussian process or neural networks to learn the system dynamics~\cite{berkenkamp2017safe, fisac2018general,shi2019neural}, and ensures robot safety by formulating an MPC problem~\cite{koller2018learning,hewing2020learning},
or constructing a barrier or Lyapunov function for safety certificates~\cite{wang2018safe,chow2019lyapunov}.
Despite being more general, these methods suffer from the curse of dimensionality. 
As we focus on the locomotion problem, we choose to leverage an approximate dynamics model widely used in locomotion control.
%

%
%
%
%
%
%

\parab{RL for Legged Locomotion.}
RL has been applied in legged locomotion to acquire complex locomotion skills~\cite{peng2020learning, kohl2004policy,tedrake2004stochastic,EndoCPG,haarnoja2018learning,yu2019sim,ha2020learning,yu2020learning}.
However, these approaches often directly learn the policy or fine-tune the learned policy in the real world without considering safety and thus require humans to recover robots when falling.
In contrast, we use the safe recovery policy to ensure safety during the learning process.
Without manually resetting robots from falls, our algorithm achieves better wall-clock time learning efficiency.

\parab{Recovery RL.}
This work~\cite{thananjeyan2021recovery} also uses a safe recovery policy to take over the control when violating safety constraints.
However, there are a few key differences.
\textbf{(1) Algorithm.} First, Recovery RL uses safety critics to predict the probability of future cost violations given the current state.
When the prediction value is above the pre-defined threshold, the safe recovery policy kicks in.
However, the safety critics are pre-trained from the policy that maximizes cost violations (\ie considering the worst-case policy) in simulation.
Such safety critics do not accurately reflect the safety performance of the learner policy when deploying in the real world.
This mismatch can potentially prevent the agent from exploring the environment (\ie too conservative), or ensuring safety (\ie too aggressive) as seen in Section \ref{sec:experiments}.
\textbf{(2) Generalization.} Second, the safety critics, which are parametrized by deep neural networks, do not guarantee correct verification of safety when visiting states unseen during training.
%
%
%
%
%
%

\section{Background}

{
\begin{figure}[t]
\centering
{
\includegraphics[width=1.0\linewidth]{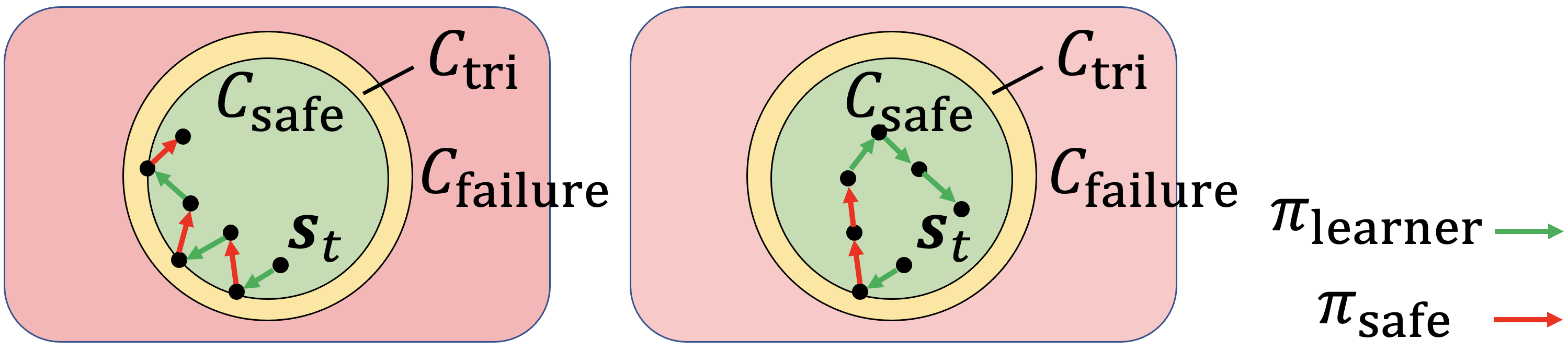}}
\vspace{-5mm}
\caption{
An illustration of safety trigger set $\mathcal{C}_{\rm tri}$ (orange), safe set $\mathcal{C}_{\rm safe}$ (green), and failure set $\mathcal{C}_{\rm failure}$ (red) with two policies $\pi_{\rm learner},$ and $\pi_{\rm safe}.$
On the left, without the reachability criteria, we encounter a frequent switch between the safe recovery policy $\pi_{\rm safe}$ and the learner policy $\pi_{\rm learner}.$ On the right, by applying the reachability criteria, we push the learning agent away from $\mathcal{C}_{\rm tri},$ which reduces the risk of violating the safety constraints.
}
\label{fig:set}
\vspace{-6mm}
\end{figure}
}

We frame our problem as a constrained Markov Decision Process (CMDP)~\cite{altman1999constrained}, defined as a tuple $<\mathcal{S},\mathcal{A},T,R,C>.$ 
Here $\mathcal{S}$ is the set of states, 
$\mathcal{A}$ is the set of actions, and
$T$ is the transition function that specifies the conditional probability $T(\vs'|\vs,\va)$ or the deterministic function $\vs'=T(\vs,\va)$ that the next state is $\vs'$ given the current state $\vs$ and action $\va.$ 
In addition, 
$R:\mathcal{S}\times \mathcal{A}\rightarrow \mathbb{R}$ is a reward function, and
$C:\mathcal{S}\rightarrow \mathbb{R}$ is a constraint cost function that indicates the failure state (\eg falling down). The reward function encodes the benefit of using action $\va$ in state $\vs,$ while the cost function encodes the corresponding constraint violation penalty (we assume the value of $C$ is non-negative).

A policy $\pi(\va|\vs)$ is a map from states to probability distributions on ${\cal A}: \va\sim\pi(\va|\vs).$ 
The state then transits from $\vs$ to $\vs'$ according to $T.$
Finally, the agent receives a reward $R(\vs,\va)$ and incurs a cost $C(\vs).$

Let $\gamma\in (0,1)$ denote a discount factor, and $\tau$ denote the trajectory $\tau = (\vs_0,\va_0,\vs_1,\cdots)$ from a policy $\pi.$
We seek a policy $\pi$ that maximizes a cumulative discounted reward:
$J_{R}(\pi)\doteq \E_{\tau\sim\pi}\left[\sum_{t=0}^{\infty}\gamma^{t} R(s_{t},a_{t})\right],\label{eq:problemFormulation_1}$
%
\setlength{\abovedisplayskip}{6pt}%
\setlength{\belowdisplayskip}{6pt}%
\setlength{\abovedisplayshortskip}{3pt}%
\setlength{\belowdisplayshortskip}{3pt}%
\setlength{\jot}{0pt}
%
%
while keeping the cumulative discounted cost below a pre-defined threshold $h:$
$J_{C}(\pi)\doteq \E_{\tau\sim\pi}\left[\sum_{t=0}^{\infty}\gamma^{t} C(s_{t})\right]\leq h.$
%
Note that we refer to systems with $J_C(\pi)>h$ as \textit{cost violations} or \textit{failure}.
One difference between conventional RL and safe RL is that we want the \textit{entire} training process to satisfy the safety constraints, not just at the end of training. 
%

\section{Algorithm}
In this section, we first provide the necessary definitions, followed by proposing the safe control approach using approximate knowledge about system dynamics. 
Finally, we conclude the section by outlining the proposed framework.

\subsection{Setup}

\begin{definition}
\label{de:1}
{\rm We define:
\begin{itemize}
\item [(1.1)] Safety trigger set $\mathcal{C}_{\rm tri}\subseteq\mathcal{S},$ a set of states that triggers $\pi_{\rm safe}.$
\item [(1.2)] Safe set $\mathcal{C}_{\rm safe}=\mathcal{C}_{\rm tri}\setminus\mathcal{S}$ in which the state does not belong to the safety trigger set;
\item [(1.3)] Failure set $\mathcal{C}_{\rm failure}=\{\vs\in\mathcal{S}:C(\vs)>0\}$ in which an non-negative cost penalty (\ie falls) incurred for certain $\vs.$
Note that $\mathcal{C}_{\rm failure}\subseteq\mathcal{C}_{\rm tri}.$
\end{itemize}
}
\end{definition}
%
%
Definition (1.3) defines unsafe states that we want to avoid.
For example, when the height of a robot is too low (\ie indication of falls), the learning agent is in $\mathcal{C}_{\rm failure}.$
Fig. \ref{fig:set} shows these three sets.
To ensure safety of the learning agent, we will also need the following assumptions.
\begin{assumption}
\label{as:1}
{\rm \ 
\begin{itemize}

\item [] An approximate knowledge of the true system dynamics $T,$ denoted by $\hat{T},$ in which their difference is bounded by ${\|T-\hat{T}\|}_2\leq\epsilon.$

\end{itemize}
}
\end{assumption}
%
%
Approximate dynamics models are often leveraged to achieve efficient planning of robot motions. For example, the centroidal dynamics model (CDM) or spring-loaded inverted pendulum model (SLIP) are commonly used to design locomotion controllers, while a unicycle model is popular for modeling wheeled robots. In this work, we utilize a centroidal dynamics model to predict the future state trajectory given the current state and $\pi_{\rm learner}$'s control inputs to certify the learning agent's safety.

\subsection{The Safe Control Approach}
Based on these definitions and assumptions, the proposed algorithm uses two policies, a \textit{learner policy} $\pi_{\rm learner},$ which maximizes the task reward, and a \textit{safe recovery policy} $\pi_{\rm safe},$ which tries to bring the agent back to safe states when reaching the safety trigger set.
%
%
A naive way to select which policy to query is as follows.
\begin{equation} \textstyle
\label{eq:control}
\va_{\rm naive}=
\begin{cases}
\pi_{\rm safe}(\cdot|\vs), & \text{if}\ \vs\in\mathcal{C}_{\rm tri} \\
\pi_{\rm learner}(\cdot|\vs), & \text{otherwise}.
\end{cases}
\end{equation}
This control approach implies that whenever the state is in the safety trigger set, the safe recovery policy takes over the control from the learner, and returns the control back when the learner is in the safe set.
However, this poses a potential issue: while in principle the control approach in Eq. (\ref{eq:control}) can ensure safety, it may cause frequent switch between $\pi_{\rm safe}$ and $\pi_{\rm learner}$, which hinders policy exploration and impacts learning efficiency.
Fig. \ref{fig:set} illustrates this observation.
To address this issue, when the learning agent is in safety trigger set, the safe recovery policy takes over the control and should return the control to the learning agent only when $\pi_{\rm learner}$ can guarantee that there would be \textit{no further switches} for a specific future horizon.
This reduces the potential use of the safe recovery policy.
Formally, we define the following planning criteria for determining when should $\pi_{\rm safe}$ hand control back to $\pi_{\rm learner}.$ 
\begin{definition}
\label{de:2}
{\rm \textbf{Reachability Planning Criteria.}

$\exists{\{\va_\tau\}}_{\tau=t}^{i-1}\in\pi_{\rm safe}$ with  $\min_{i\in[t+1,t+w-1]} i$ such that $\{\va_{\tau}\}_{\tau=i}^{t+w-1}\in\pi_{\rm learner}$ with $\vs_\tau\notin\mathcal{C}_{\rm tri}$ where $w\in\R^{+}$ is the planning step.
}
\end{definition}
Definition~(\ref{de:2}) says that we want to find a minimal step $i$ (\ie the minimal intervention) such that all the remaining actions are from $\pi_{\rm learner}$ after a minimal initial sequence of actions from $\pi_{\rm safe}.$
This ensures that when $\pi_{\rm safe}$ hands the control back to $\pi_{\rm learner},$ $\pi_{\rm learner}$ can maximally reduce future use of the safe recovery policy.
However, Definition (\ref{de:2}) is computationally intensive and non-convex, which can limit the practicality of the proposed approach for real-time validation of safety in some applications such as robots with onboard processing. 
An alternative is to remove the planning component in reachability planning criteria and only verify whether future states are in the safety trigger set given all actions from $\pi_{\rm learner}.$
Formally, $\pi_{\rm safe}$ returns the control back to $\pi_{\rm learner}$ if the following criteria are satisfied.
\begin{definition}
\label{de:3}
{\rm \textbf{Reachability Criteria.}

Check ${\{\va_\tau\}}_{\tau=t}^{w+t-1}\in\pi_{\rm learner}$  such that $\{\vs_\tau\}_{\tau=t+1}^{w+t}\notin \mathcal{C}_{\rm tri}.$
}
\end{definition}
The criteria say that at time $t,$ we check whether the future states are in $\mathcal{C}_{\rm tri}$ for $w$ steps.
Hence, the final proposed safe control approach becomes
\begin{equation}\textstyle
\label{eq:control_fianl}
\va_t=
\begin{cases}
\pi_{\rm safe}(\cdot|\vs_{t}), &\underbrace{\text{if}\ (\vs_t\in\mathcal{C}_{\rm tri})}_{\pi_{\rm safe}~\text{trigger criterion}}~\lor~\;\;\\
&\big((\va_{t-1}\in\pi_{\rm safe})~\land~\;\;\\
& (\nexists{\{\va_\tau\}}_{\tau=t}^{w+t-1}\in\pi_{\rm learner}~\text{s.t.}~\;\;\\ & \underbrace{\{\vs_\tau\}_{\tau=t+1}^{w+t}\notin \mathcal{C}_{\rm tri})\big)}_{\pi_{\rm safe}~\text{handing control back criterion}} \\
\pi_{\rm learner}(\cdot|\vs), & \text{otherwise}.
\end{cases}
\end{equation}
There are two conditions that we use a safe recovery policy.
The first condition (\ie safe recovery policy trigger criterion) is when the current state is in the safety trigger set (\ie $\vs_t\in\mathcal{C}_{\rm tri}$). 
The second condition (\ie safe recovery policy handing control back criterion) is that when the previous action is from safe recovery policy and there \textit{does not}
exist a sequence of actions produced by the learner policy such that the resulting states are not in the safety trigger set (\ie $(\va_{t-1}\in\pi_{\rm safe})~\land~(\nexists{\{\va_\tau\}}_{\tau=t}^{w+t-1}\in\pi_{\rm learner}~\text{s.t.}~\{\vs_\tau\}_{\tau=t+1}^{w+t}\notin \mathcal{C}_{\rm tri})$).
Note that the reachability criteria (\ie the second condition) are only checked if the previous action is from $\pi_{\rm safe},$ which determines when to \textit{hand the control back} to the learner.
This is different from Recovery RL~\cite{thananjeyan2021recovery}, in which $\pi_{\rm safe}$ hands the control back to the learner whenever the prediction of future cost constraint violations are below a certain threshold at any state in the environment.
Our approach allows the agent to have enough freedom to explore the states that are near $\mathcal{C}_{\rm tri}$ while avoiding constant use of $\pi_{\rm safe}.$
$\pi_{\rm safe}$ is triggered only when needed.
Finally, the approximate dynamics $\hat{T}$ is used to unroll the future states given the current state and $\pi_{\rm learner}$ for the reachability criteria. 

\begin{algorithm}[t]
\centering
\caption{Safe RL with Approximate System Dynamics}
\label{algo:proposed_algo}
\begin{algorithmic}[1]
    \State \textbf{Given:} \textbf{(1)} Knowledge about system dynamics $\vs'=\hat{T}(\vs,\va)$ or $\vs'=\hat{T}(\cdot|\vs,\va),$
    \textbf{(2)} an initial learner policy $\pi^0_{\rm learner}=\pi(\cdot|\vtheta^0_{\rm learner})$ parameterized by $\vtheta^0_{\rm learner},$ 
    \textbf{(3)} a safe recovery policy $\pi_{\rm safe}(\cdot|\vs),$ 
    \textbf{(4)} safety trigger set $\mathcal{C}_{\rm tri}$, and 
    \textbf{(5)} a trajectory buffer $\mathcal{B}$ 
    \State \textbf{Parameter:} \textbf{(1)} the number of policy updates $K,$ 
    \textbf{(2)} the number of interactions  $W,$
    \textbf{(3)} planning horizon $w$
    \For{$k=0,1,\cdots,K$}
        \For{$t=1,\cdots,W$}
            \State Use control approach in Eq. (\ref{eq:control_fianl}) 
            
            
            \State Observe $\vs_{t+1}, r_t=R(\vs_t,\va_t)$
            \State Store data $(\vs_t, \va_t^{\rm learner}, r_t-z)$ in $\mathcal{B},$ where
            
            ~~~$z=1$ if $\va_t\in\pi_{\rm safe},$ and zero otherwise.
        \EndFor
        \State Update $\vtheta_{\rm learner}^k$ using $\mathcal{B}$
    \EndFor
\end{algorithmic} 
\end{algorithm}

\subsection{Reinforcement Learning with the Safe Control Approach}
Algorithm \ref{algo:proposed_algo} illustrates the proposed framework with the safe control approach.
In addition, when storing the data trajectory in the replay buffer (line 7), we always use the action proposed by $\pi_{\rm learner}$ with the negative penalty for indicating the use of $\pi_{\rm safe}.$
This implies that we treat $\pi_{\rm safe}$ as \textit{part of the environment}, and we want the learning agent to reduce the use of $\pi_{\rm safe}$ over the learning process to embed safety to the agent.
Note that one can also terminate the episode when triggering $\pi_{\rm safe}$ to discourage the use of $\pi_{\rm safe}.$
\section{Theoretical Analysis for Dynamics Error}
The proposed reachability criteria rely on the knowledge about the system dynamics $T$ to predict the future state trajectory.
Hence, we want to quantify the effect of the approximation error of the system dynamics on the performance of the task objective.
Let $\{\vy_t\}\in\mathcal{S}$ denote a target safe trajectory that the learning agent wants to track.
Then, we consider the following simplified problem in which we want to find a sequence of actions that minimize the tracking errors for $W$ steps between the safe and the planned trajectories under the proposed control approach.
\begin{align}
    \argmin_{\va_{1}\cdots\va_{W}}\sum_{t=1}^{W}\underbrace{{\|\hat{T}(\vx_{t},\va_{t})-\vy_{t+1}\|}_2}_\text{task objective}~\text{s.t.}~\text{control method in}~(\ref{eq:control_fianl})
    \label{eq:problem}
\end{align}

\begin{assumption}
\label{as:2}
{\rm We assume:
\begin{itemize}
\item [(2.1)] \textbf{Dynamics Parameterization}: The true dynamics $\vs'=T(\vs,\va)$ is a time-invariant linear dynamics $\vs'=\mA\vs+\mB\va,$ where $\mA$ and $\mB$ are system matrices.
\item [(2.2)] \textbf{Approximate Dynamics}: The dynamics $\hat{T}(\vs,\va)$ is a time-invariant linear dynamics $\vs'=\hat{\mA}\vs+\hat{\mB}\va.$
\item[(2.3)] \textbf{Smoothness}: The system dynamics $T(\vs,\va)$ is $\alpha$-Holder continuous in the second-order argument for $\alpha\in\R^{+},$ \ie for all $\vs,$ there exists $G\in\R^{+},$ such that
\begin{align} \textstyle
    {\|T(\vs,\va_1)-T(\vs,\va_2)\|}_2\leq G{\|\vs_1-\vs_2\|}_2^{\alpha}, \forall \vs_1,\vs_2.\nonumber
\end{align}
$G$ and $\alpha$ control the sensitivity of the system dynamics function to a perturbation in $\va.$
This is also true for $\hat{T}.$
\item [(2.4)] \textbf{Size of the Control Space}: For a constant $D,$
$\diam({\mathcal A}) \leq D,$ where ${\mathcal A}$ is the action space and $\diam$ is the diameter of the set (\ie ${\|\va_1-\va_2\|}_2\leq D,$ for every $\va_1, \va_2\in\mathcal{A}$).
\end{itemize}
}
\end{assumption}
Asm. (2.1) and (2.2) define the structure of the system dynamics, which has been applied to a wide range of tasks such as legged locomotion~\cite{di2018dynamic}. 
In addition, Asm. (2.3) says that the function value does not change abruptly given two points.
Asm. (2.4) implies that the size of $\mathcal{A}$ is bounded.

To provide a performance guarantee of Problem (\ref{eq:problem}), we compare the value of the task objective of the proposed algorithm running \textit{online} (denoted by ${\rm cost}(\hat{T})$) against
the optimal \textit{offline} algorithm, which makes the optimal decision with full knowledge of the system dynamics $T$ (denoted by ${\rm cost}(T)$).
Hence, we have the following Theorem.
\begin{theorem}
\label{thm:tracking_error}
\textbf{(Performance Guarantee.)}
We have 
\begin{align}\textstyle
    \E[{\rm cost}(\hat{T})]\leq&\E[{\rm cost}(T)] + 2^{\frac{\alpha}{2}+1}GW\cdot{|\sigma_{1}(\mA-{\hat{\mA}})|}^{\alpha}\nonumber\\
    &\cdot\bigg({|\sigma_{1}(\mB-{\hat{\mB}})|}^{\alpha} a_{\rm max}^{\alpha}\Big(\frac{2}{\alpha+2}\Big)+{\|\vs_1\|}_2^{\alpha}
    \bigg),\nonumber
\end{align}
where
$\sigma_1(\cdot)$ is the largest singular value of a matrix,
$\vs_1$ is the initial state, and
scalar value $a_{\rm max}:=\max_{t\in\{1,\cdots,W\}}{\|\va_t\|}_2.$
\end{theorem}

\textit{Proof:} \ 
Please check our project website for a proof.

%
We now make several observations for Theorem \ref{thm:tracking_error}.\\
\textbf{(1)} If there is no dynamics error, \ie $\mA=\hat{\mA}, \mB=\hat{\mB},$ then $\E[{\rm cost}(\hat{T})]\leq\E[{\rm cost}(T)],$ which implies that the online decision planning performs as good as the offline algorithm.
However, when there is a dynamics error,
the leading singular value of the system matrices difference will potentially lower the performance of the task objective with the cost violations.
\\
\textbf{(2)} If the system dynamics is less smoother, \ie $\alpha$ is large, then the performance of the proposed algorithm degrades.
This implies that when the system is unstable and unpredictable, we have a large tracking error.
\\
\textbf{(3)} It shows the performance drop to the offline algorithm grows linearly w.r.t the horizon $W.$
In addition, the smoothness value $G$ represents how \textit{predicable} the dynamics is--the smaller the value is, the more predicable the dynamics are.\\
\textbf{(4)} In practice for the locomotion tasks in the paper, their system dynamics are non-linear. However, we can treat it \textit{locally} linear when the time step is small enough. We leave the non-linear version of the bound as the future work.
In summary, Thm \ref{thm:tracking_error} provides an intuition about the effect of horizon $W$ and model errors.

\section{Experiments}
\label{sec:experiments}
We study the following questions: 
\textbf{(1)} How does our algorithm perform compared to other baselines in terms of constraint satisfaction and final reward?
\textbf{(2)} What is the effect of the planning horizon and the safety trigger set?
%

\subsection{Setup}
\parab{Robots.}
We evaluate our approach on a simulated Unitree Laikago \cite{Laikago}, a quadrupedal robot that weighs 24kg and has 12 actuated joints.
For the real-world experiment, we use a Unitree A1 \cite{A1}, a quadrupedal robot that weighs 12.7kg with better hardware reliability and agility.

\parab{Hierarchical Policies.}
We use a hierarchical policy framework that combines RL and optimal control for $\pi_{\rm learner}$ and $\pi_{\rm safe}.$ 
This framework consists of a high-level RL policy that produces gait parameters and feet placements, and a low-level model predictive control (MPC) controller \cite{di2018dynamic} that takes in these parameters to compute desired torque for each motor in the robot.
Instead of directly commanding the motor's angle, this approach offers stable operation and streamlines the policy training due to a smaller action space and a robust MPC controller.
The gait parameters include stepping frequency ($\Omega\in\R$, the frequency of completing one swing-stance cycle of each leg), swing ratio ($p_{\rm swing}\in\R$, the portion of swing duration), and phase offsets relative to the front-right leg ($\theta_{1},\theta_{2},\theta_{3}\in\R$, the angle added to the front-right leg's phase angle for the other three leg's phase angles).
The feet placements include $p_{{\rm FR},y}, p_{{\rm FL},y}, p_{{\rm RR},y}$ and $p_{{\rm RL},y}\in\R,$ which are the desired feet positions in the side-way (y) direction (FR: front-right; FL: front-left; RR: rear-right; RL: rear-left).
Such policy parameterization allows us to take advantage of the MPC controller while being expressive enough to produce a diverse set of gaits.
The MPC controller uses a \textit{centroidal dynamics model} (CDM) proposed in~\cite{di2018dynamic} to serve as $\hat{T}$.
The dynamics can be expressed as a linear dynamic model $\vs'=\mA\vs+\mB\va,$ where $\mA$ and $\mB$ are state and control matrices that depend on the gait parameters and feet positions.
When rolling out the states using $\hat{T},$ we use the action from $\pi_{\rm learner}$ and keep the other MPC parameters fixed.
The state in CDM is
$\vs_t={[x,y,z,\dot{x},\dot{y},\dot{z},\phi,\theta,\psi,\dot{\phi},\dot{\theta},\dot{\psi}]}^T,$ where $x,y,z\in\R$ are the robot's position, $\dot{x},\dot{y},\dot{z}\in\R$ are the velocity, $\phi,\theta,\psi$ are the roll, pitch, yaw, and $\dot{\phi},\dot{\theta},\dot{\psi}$ are the angular velocity.\\
%
\parab{Tasks.}
We compare our algorithm with existing approaches on four legged locomotion tasks shown in Fig.~\ref{fig:algorithm} and Fig.~\ref{fig:main_result}. \\
\textbf{(1) Efficient Gait.} The robot learns how to walk with low energy consumption.
The robot is rewarded for consuming less energy.
The actions of the policy network produce the \textit{delta change} of the gait parameters at each step, including the delta change of the stepping frequency $\Omega$, the swing ratio $p_{\rm swing}$, and the phase offsets $\theta_1, \theta_2, \theta_3.$\\
\textbf{(2) Catwalk.} The robot learns a catwalk gait pattern, in which left and right two feet are close to each other.
This is challenging because by narrowing the support polygon, the robot becomes less stable.
The robot is rewarded for using narrow foot placement: $R:=e-{(p_{{\rm FR},y}-p_{{\rm FL},y})}^2-{(p_{{\rm RR},y}-p_{{\rm RL},y})}^2,$ where $e\in\R^{+}$ is a positive survival bonus to make the reward non-negative.
The actions of the policy network produce the gait parameters in the efficient gait task and the feet placement $p_{{\rm FR},y}, p_{{\rm FL},y}, p_{{\rm RR},y}$ and $p_{{\rm RL},y}.$
\\
\textbf{(3) Two-leg Balance.} The robot learns a two-leg balance policy, in which the front-right and rear-left feet are in stance, and the other two are lifted.
The robot can easily fall without delicate balance control because the contact polygon degenerates into a line segment.
The robot can only control these two stance feet, and is rewarded for staying at a target height of 0.45m:
$R:=e-{(z-0.45)}^2.$
The action includes the delta change of the stepping frequency $\Omega$, the swing ratio $p_{\rm swing}$, and the phase offset $\theta_3.$
\\
\textbf{(4) Pacing.} We want to produce a desired stepping frequency and the swing ratio for performing the pacing behavior under different desired speeds.
The robot is rewarded for matching target speed: $R:=e-{\|{[\dot{x},\dot{y},\dot{z}]}^T-{[\dot{x}_{\rm target},\dot{y}_{\rm target},\dot{z}_{\rm target}]}^T\|}^2_2,$
where $\dot{x}_{\rm target}, \dot{y}_{\rm target}, \dot{z}_{\rm target}$ are the target velocities.
The action includes the delta change of the stepping frequency $\Omega$, and the swing ratio $p_{\rm swing}.$
Note that here we fixed the phase offsets to be $\theta_1=\pi, \theta_2=0, \theta_3=\pi$ to enforce the pacing gait: legs on the same side move in sync.
{
\addtolength{\tabcolsep}{-4pt}
\begin{figure*}[t]
\centering
{
\begin{tabular}[b]{@{}c@{}}%
\textbf{Catwalk}\\
\includegraphics[width=0.25\linewidth]{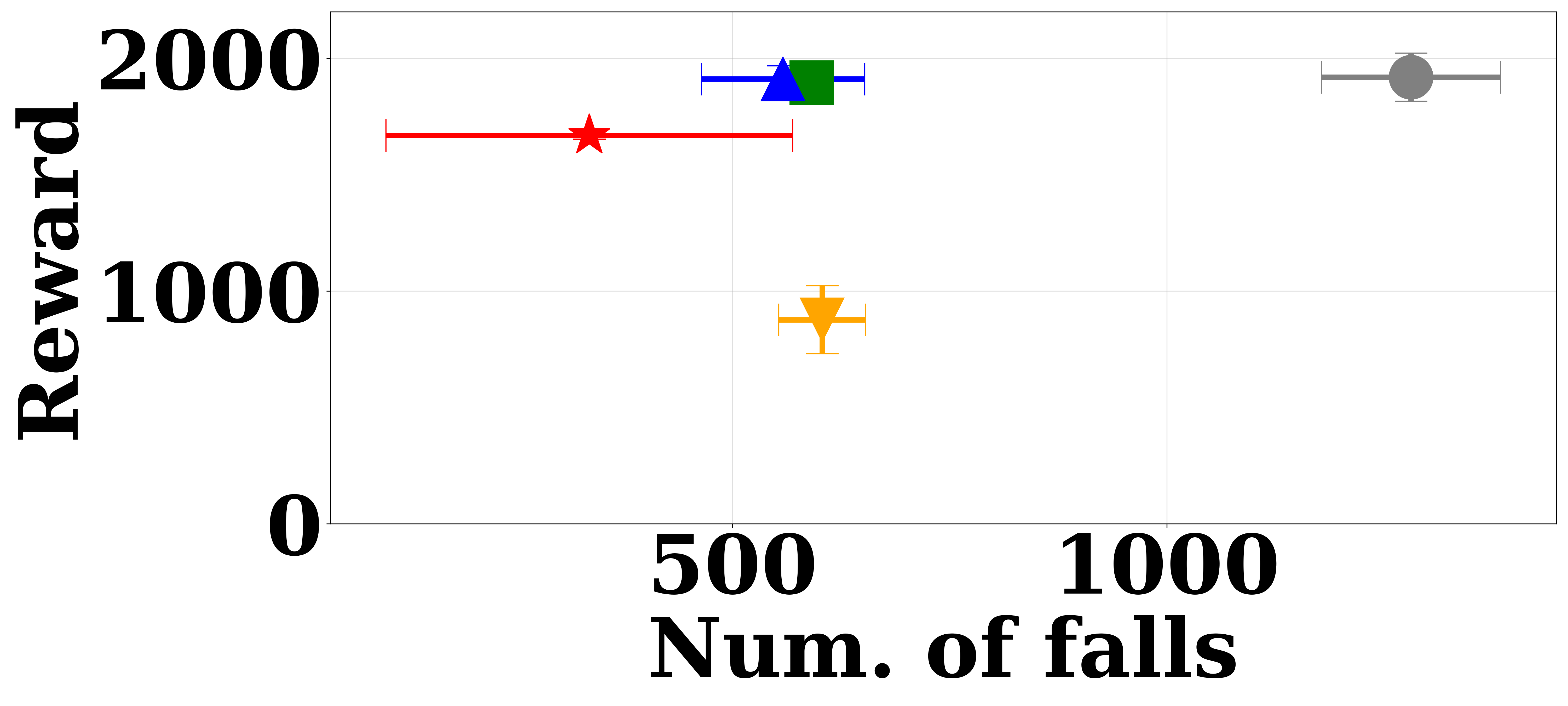}\hspace{-1mm}
\includegraphics[width=0.07\linewidth]{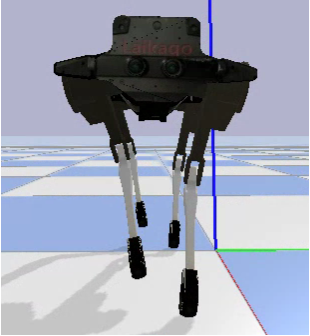}
\end{tabular}
\begin{tabular}[b]{|@{}c@{}|}%
\textbf{Two-leg balance}\\
\includegraphics[width=0.25\linewidth]{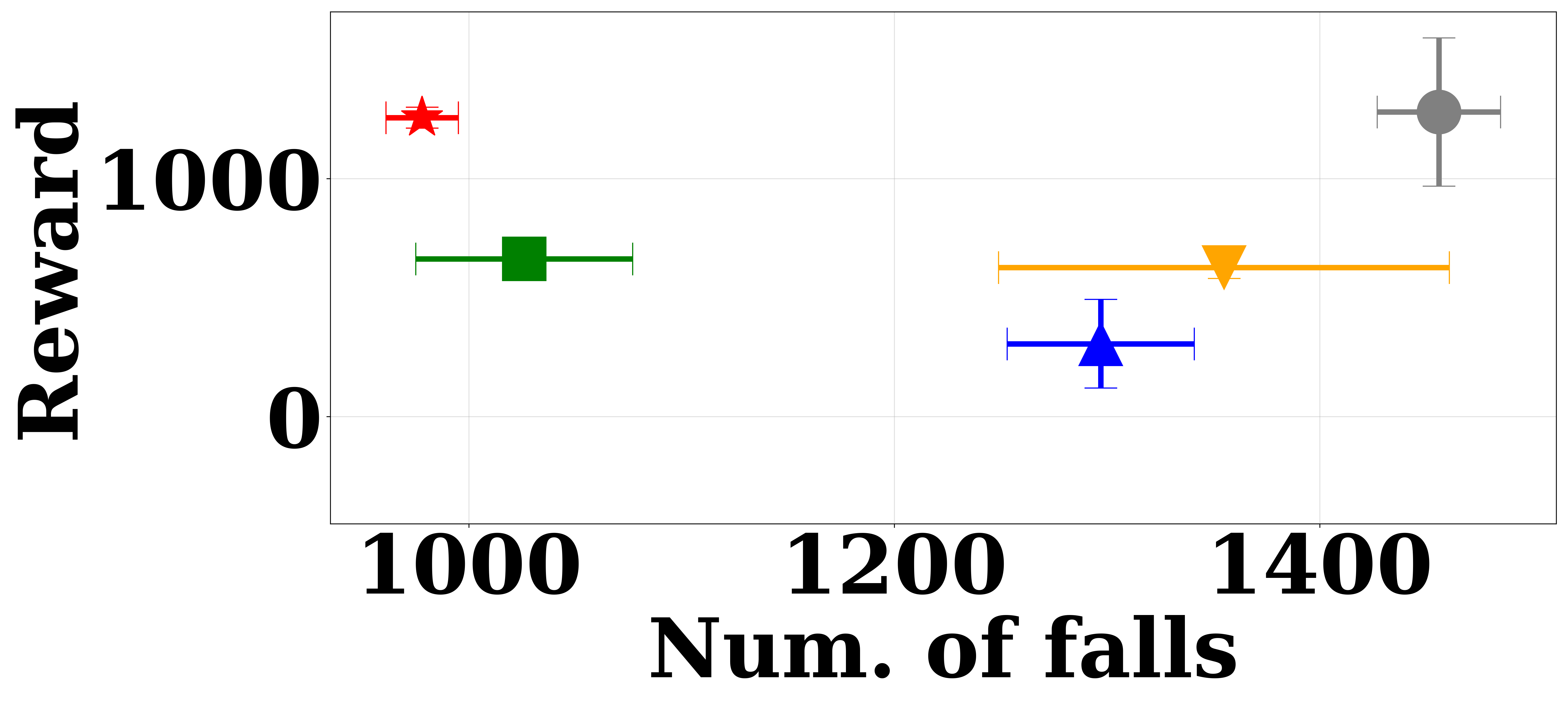}\hspace{-1mm}
\includegraphics[width=0.07\linewidth]{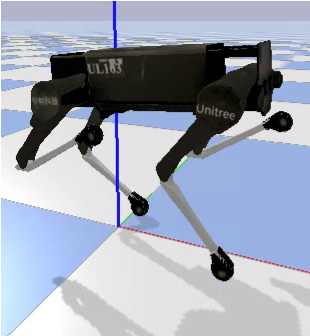}
\end{tabular}
\begin{tabular}[b]{@{}c@{}}
\textbf{Pacing}\\
\includegraphics[width=0.25\linewidth]{
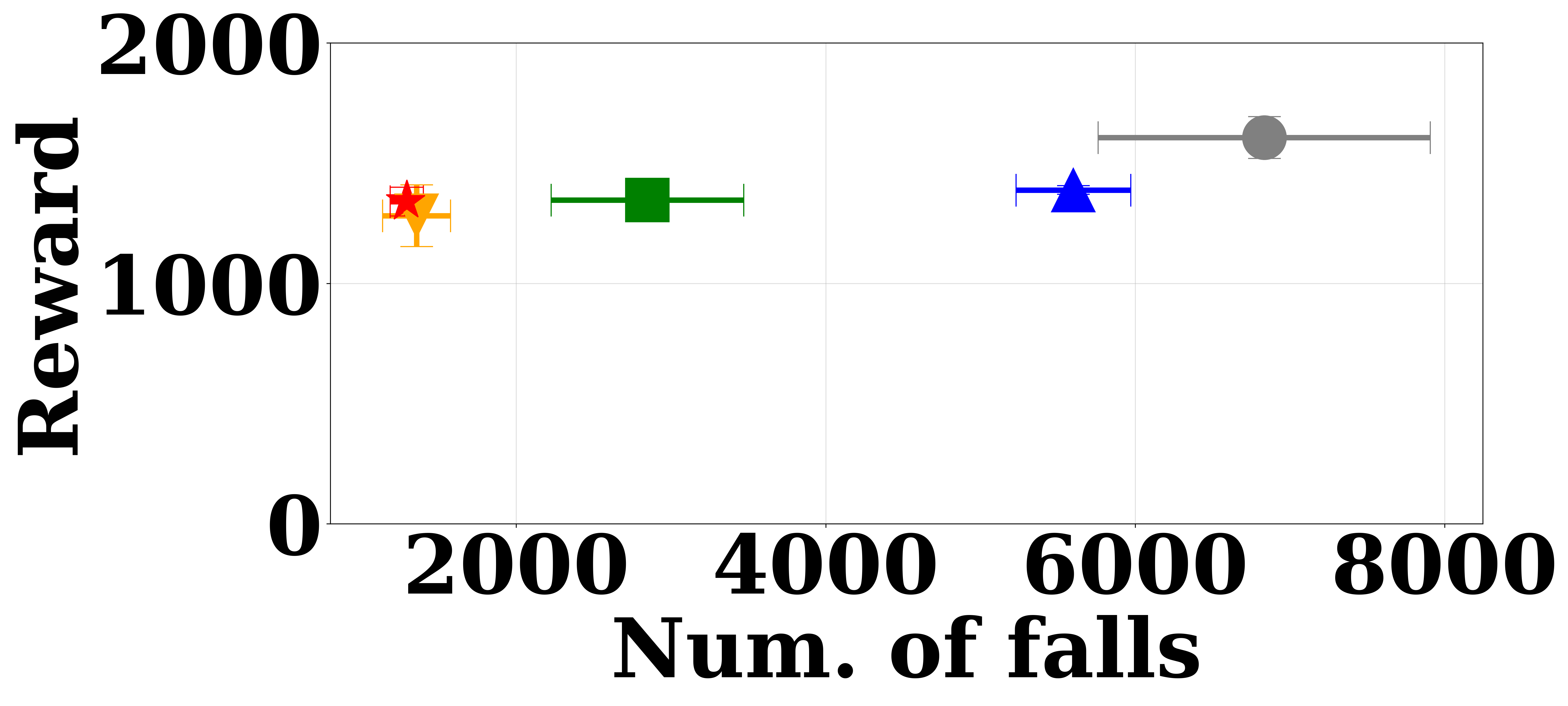}\hspace{-1mm}
\includegraphics[width=0.07\linewidth]{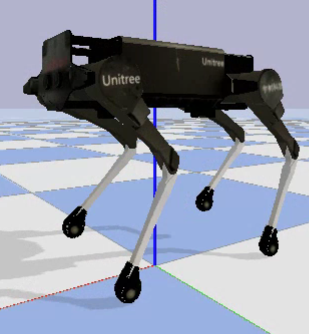}
\end{tabular}

\includegraphics[width=0.625\linewidth]{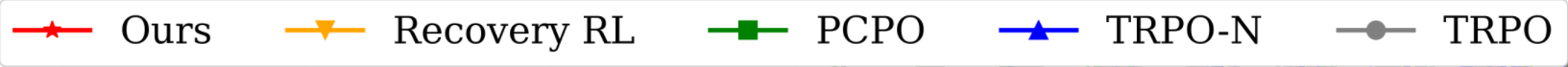}

}
\vspace{-1mm}
\caption{
\textbf{Simulation Results.} We report the total number of falls versus final rewards for the tested algorithms and task pairs over five runs.
We observe that the proposed approach achieves the fewest number of falls while having comparable reward performance (\ie in the top-left corner) (best viewed in color).
}
\label{fig:main_result}
\vspace{-7mm}
\end{figure*}
}
\begin{figure*}[t]
\centering
\textbf{(a)}
\setlength\tabcolsep{2.0pt}{\scalebox{0.87}{
\begin{tabular}{l*{3}{c}r}
\\ 
& Num. of falls & Num. of $\pi_{\rm safe}$ & Reward  \\ \hline
Ours $w=0$   & 551 $\pm$ 84 & 106,280 $\pm$ 4,921 & 1517 $\pm$ 282 \\
Ours $w=5$    & 548 $\pm$ 87 & \textbf{104,532 $\pm$ 15,496}& \textbf{1677 $\pm$ 32}\\
Ours $w=10$  & 335 $\pm$ 34 & 116,155 $\pm$ 55,715 & 1575 $\pm$ 275 \\
Ours $w=15$  & 299 $\pm$ 86 & 157,285 $\pm$ 5,085 & 1506 $\pm$ 229  \\
Ours $w=20$  & \textbf{204 $\pm$ 20} & 194,198 $\pm$ 5,114 & 1433 $\pm$ 307 \\
Recovery RL  & 603 $\pm$ 50 & 485,633 $\pm$ 325,133 & 894 $\pm$ 163   \\
\bottomrule
\end{tabular}}}
\textbf{(b)}
\setlength\tabcolsep{2.0pt}{\scalebox{0.87}{
\begin{tabular}{l*{3}{c}r}
\\ 
& Num. of falls & Num. of $\pi_{\rm safe}$ & Reward  \\ \hline
Ours $\mathcal{C}_{\rm tri}^1$ & 335 $\pm$ 234 & \textbf{116,155 $\pm$ 55,715} & \textbf{1575 $\pm$ 275} \\
Ours $\mathcal{C}_{\rm tri}^2$ & 313 $\pm$ 32 & 141,488 $\pm$ 16,536 & 1525 $\pm$ 180   \\
Ours $\mathcal{C}_{\rm tri}^3$ & \textbf{275 $\pm$ 137} & 246,651 $\pm$ 25,982 & 677 $\pm$ 154   \\
Recovery RL  & 603 $\pm$ 50 & 485,633 $\pm$ 325,133 & 894 $\pm$ 163   \\
\bottomrule
\end{tabular}}}
\vspace{-1mm}
\caption{
\textbf{Simulation Results.}
\textbf{(a)} Ablations on the planning horizon $w$ in the catwalk task in terms of the number of falls, the number of uses of $\pi_{\rm safe}$, and the reward performance.
Results suggest that with an increasing number of $w,$ one can reduce the number of falls with more safety trigger events.
\textbf{(b)} Ablations on the design of the safety trigger set $\mathcal{C}_{\rm tri}$ in the catwalk task.
Results suggest that the looser $\mathcal{C}_{\rm tri}$ is, the fewer number of falls are. 
The best numbers are bold.
}
\vspace{-4mm}
\label{fig:ablation}
\end{figure*}
\\
\parab{Baselines.}
The goal is to demonstrate that the proposed approach can enable safe training using the knowledge of the system dynamics to determine switching timing.
In addition, we position our paper in the field of RL.
Hence we compare our method with the following state-of-the-art RL or safe RL baselines. \\
\textbf{(1) TRPO.} Trust region policy optimization (TRPO~\cite{schulman2015trust}) only optimizes the reward objective. 
This will serve as a cost constraint performance \textit{lower} bound. \\
\textbf{(2) PCPO.} Projection-based constrained policy optimization (PCPO, \cite{yang2020projection}) optimizes the reward while using projection to satisfy the cost constraint.
In addition, PCPO is the state-of-the-art model-free approach over CPO \cite{achiam2017constrained}.\\
\textbf{(3) TRPO w/ negative penalty (TRPO-N).} It optimizes the reward objective with an added penalty for entering $\mathcal{C}_{\rm failure}:$ $R(s,a)-C(s).$ This is to show that the penalty-based approach would fail when the dynamics are complex.\\
\textbf{(4) Recovery RL.} We are inspired by Recovery RL \cite{thananjeyan2021recovery}, which also consists of two policies: a safe recovery policy and a learner policy.
%
%
This is a direct comparison of the proposed switch criteria with the value function-based method. 
We use hierarchical policies for all the baselines.\\
\parab{Implementation Details.}
We implement these tasks using Pybullet~\cite{coumans2017pybullet} simulator.
We use a two-layer multi-layer perceptron (MLP)-based policy network with a tanh activation function for $\pi_{\rm learner}$ and $\pi_{\rm safe}.$
The low-level MPC controller runs at the frequency of 250Hz, and the policy network predicts the gait parameters with 125Hz.\\
\textbf{(1) Observation Space.}
The observation space of the policy network includes the previous gait parameters, the height of the robot, base orientation, linear and angular velocities.\\
\textbf{(2) Safe Recovery Policy $\pi_{\rm safe}$.} We want to learn a policy that can safely bring the agent to a balanced and stationary state.
We use the above hierarchical approach to train the policy, and add random perturbation to the robot to simulate the unstable and unsafe behavior encountered during the learning process. 
This allows us to train a policy that can stabilize the robots robustly.
The robot is rewarded for being stable with zero velocity and maintaining the desired height:
$R:=e-{\|{[z,\dot{x},\dot{y},\dot{z},\phi,\theta,\psi,\dot{\phi},\dot{\theta},\dot{\psi}]}^T-{[0.45, 0, 0, 0, 0, 0, 0,0,0,0]}^T\|}^2_2.$
The action includes the gait parameters used in the efficient gait.
Note that all $\pi_{\rm safe}$ in the simulation are learned first before learning $\pi_{\rm learner}$ in the proposed method and Recovery RL.
We stop training $\pi_{\rm safe}$ during learning $\pi_{\rm learner}.$
For the real-world experiments, we compare two types of $\pi_{\rm safe}$: a learned MLP-based $\pi_{\rm safe}$ and a simulation-tuned MPC controller. We find that the optimized MPC controller works better in the real world, possibly because the smaller number of tunable parameters leads to a smaller sim-to-real gap. As such, we use this controller in our real-world experiments.\\ 
\textbf{(3) Constraint Cost Function $C$.} For all the tasks, $C(\cdot)$ outputs $1$ when the height is below 0.1m and zero otherwise. \\
\textbf{(4) Safety Trigger Set $\mathcal{C}_{\rm tri}$.}
To ensure safety, we want the robot to maintain a certain height, stay upright with smaller side-tilting velocities.
Hence we use the following safety trigger set for Laikago:
$\mathcal{C}_{\rm tri}=\{\vs\in\mathcal{S}: z<0.4 \lor z>0.55 \lor |\phi|>0.26 \lor |\theta|>0.26 \lor |\dot{y}|>0.5 \lor |\dot{\phi}|>0.5\},$ where the unit for angle is in radian. For A1 robot, we use the same safety trigger set except for the height bounds to be $z<0.2$ and $z>0.3$ since the robot is smaller.
We design our $\mathcal{C}_{\rm tri}$ inspired by the capturability theory~\cite{chen2022quadruped}. In particular, we choose an initial tight $\mathcal{C}_{\rm tri}$ such that the corresponding capture point for the robot does not exceed the task space (0.15m for Laikago and 0.1m for A1). We then fine-tune $\mathcal{C}_{\rm tri}$ on the real A1 robot with a random policy to identify $\mathcal{C}_{\rm tri}$ that can prevent the robot from falling. 
%
In addition, we also conduct an ablation on the effect of the safety trigger set where we use a \textit{stricter} criterion:
$\mathcal{C}_{\rm tri}^{2}=\{\vs\in\mathcal{S}: z<0.4, z>0.55, |\phi|>0.26, |\theta|>0.26, |\dot{y}|>0.375, |\dot{\phi}|>0.375\},$ and\\
$\mathcal{C}_{\rm tri}^{3}=\{\vs\in\mathcal{S}: z<0.4, z>0.55, |\phi|>0.26, |\theta|>0.26, |\dot{y}|>0.25, |\dot{\phi}|>0.25\}.$ 
Note that $\mathcal{C}_{\rm tri}^{1}:=\mathcal{C}_{\rm tri}$ (default criteria).
\\
\textbf{(5) Planning Horizon $w$.} We also ablate the planning horizon $w$ (one step is 0.016 seconds) to see its effect on the number of uses of the safe recovery policy and falling events.
Specifically, we use $w=0, 5, 10, 15, 20.$
Other than explicitly specified, we use $w=10$ as the default in simulation. 
For the real-world experiment, we use $w=0$ for $\pi_{\rm safe}$ that lasts 1 second due to computation budget in planning and we find this approach empirically works well.

{
\addtolength{\tabcolsep}{-4pt}
\begin{figure*}[t]
\centering
{
\begin{tabular}[b]{@{}c@{}}%
\textbf{Efficient gait}\\
\includegraphics[width=0.27\linewidth]{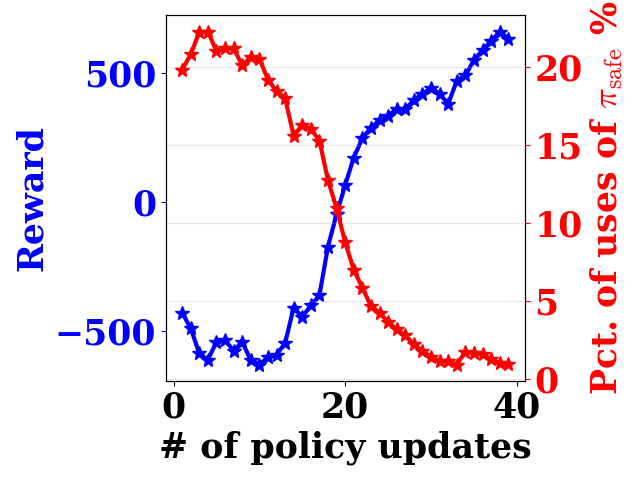}
\end{tabular}
\begin{tabular}[b]{|@{}c@{}}%
\textbf{Catwalk}\\
\includegraphics[width=0.27\linewidth]{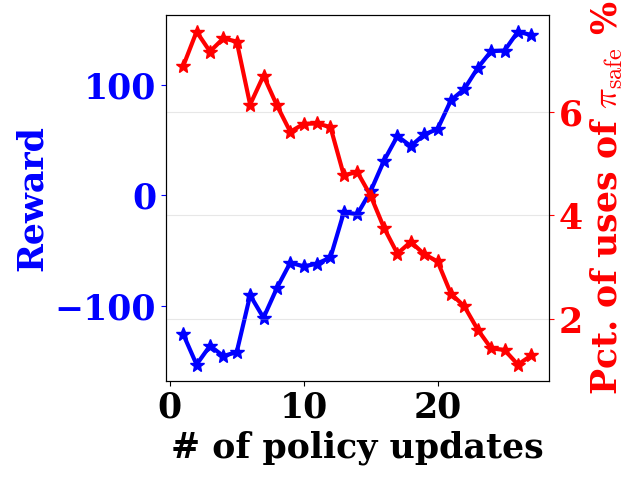}
\end{tabular}
\begin{tabular}[b]{|@{}c@{}}%
\textbf{Two-leg balance}\\
\includegraphics[width=0.27\linewidth]{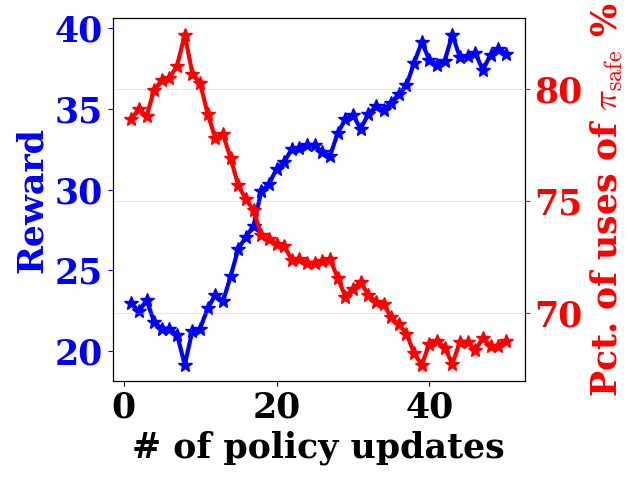}
\end{tabular}
}
\vspace{-4mm}
\caption{
\textbf{Real-world Experiments.}
We report the real-world experiment results of reward learning curves and the percentage of uses of $\pi_{\rm safe}$ on the efficient gait, catwalk, and two-leg balance tasks.
We observe that our algorithm is able to improve the reward while avoiding triggering $\pi_{\rm safe}$ over the learning process.
The learned policies are shown in Fig.~\ref{fig:algorithm}.
}
\label{fig:a1_learning}
\vspace{-5mm}
\end{figure*}
}

\subsection{Results}
\parab{Simulation Results.}
The total number of falls versus the final reward value are shown for all tested algorithms and tasks in Fig. \ref{fig:main_result}.
%
%
The learning performance for baseline oracle, TRPO, indicates the reward and the constraint performance when the safety constraint is \textit{ignored}.
Ideally, we want the algorithm to be in the \textit{top-left} corner (more rewards and fewer falls).
Overall, we find that our algorithm is able to improve the reward while achieving the fewest number of falls in all tasks (in the top-left corner).
In addition, we observe that 
\textbf{(1)} Recovery RL has more falls and cannot improve reward effectively in the catwalk tasks.
This is because the safety critic cannot predict the cost violation well;
\textbf{(2)} TRPO-N requires a significant effort to select a good value of $C$ to balance between the reward improvement and the constraint satisfaction;
\textbf{(3)} PCPO has the moderate reward and constraint performance due to the projection step.
We also observe that the learning algorithm in the two-leg balance (7.19\% for our algorithm of the total of 13,600 training trajectories) and pacing tasks (9.50\%) tend to have more falls than that of the catwalk task (2.46\%).
This is because the robot with only two stance legs is highly unstable, and the pacing gait is different from the gait used in the safe recovery policy, which creates a sudden change of the gait and hence causes a wiggling movement. \\
%
%
\parab{Ablation Studies.}
To demonstrate the efficacy of reachability criteria in the proposed framework, we ablate the planning horizon $w.$ 
Fig. \ref{fig:ablation}(a) shows the results with $w=0, 5, 10, 15, 20$ in the catwalk task in terms of the total number of falls, the number of uses of $\pi_{\rm safe},$ and the reward.
Note that when $w=0,$ it is equivalent to the removal of the reachability criteria in our algorithm, and similar to a shielding-based MPC approach in~\cite{bastani2021safe}.
In addition, $w=0$ does not mean our method is similar to recovery RL since it can trigger $\pi_{\rm safe}$ before reaching the safety boundary.
Overall, we observe that our approach outperforms Recovery RL in terms of the number of uses of $\pi_{\rm safe}$ and the reward, no matter how long the horizon is.
This is because the safety critic in Recovery RL is too conservative in predicting the cost violation, which hinders the exploration of the agent (reducing the reward).
In addition, the safety critic is pre-trained based on the worst-case policy that maximizes the cost violation.
This leads to a conservative learning strategy.
%
%
Furthermore, as we keep increasing $w$ to 20, the number of falls reduces with more uses of $\pi_{\rm safe}$ and fewer rewards.
This is expected since $\pi_{\rm safe}$ intervenes the learning agent more often, which reduces the reward performance.
This shows a trade-off between safety and reward performance, and the horizon $w$ gives users a knob to tune based on the problems.
%
%
%
%

%
Furthermore, Fig. \ref{fig:ablation}(b) ablates the design of the safety trigger set $\mathcal{C}_{\rm tri}$ in the catwalk task.
We see that the stricter $\mathcal{C}_{\rm tri}$ is, the fewer the falls and the more uses of $\pi_{\rm safe}$ are.
%
%
This observation implies one needs to design a good $\mathcal{C}_{\rm tri}$ based on the recovery capability of $\pi_{\rm safe}.$\\
\parab{Real-world Experiments.}
%
%
Fig. \ref{fig:a1_learning} shows the reward and the percentage of uses of $\pi_{\rm safe}$ among the trajectory steps collected for one policy update on the efficient gait, catwalk, and two-leg balance tasks.
For the efficient gait and the catwalk tasks, one policy update corresponds to 10 trajectories of a total of 4,000 steps, and for the two-leg balance task, it corresponds to 5 trajectories of a total of 2,000 steps.
Here, all the policies are trained from scratch, except for the two-leg balance task with the policy pre-trained in the simulation since it requires more samples to train.
Note that we do not include the other baselines since they cannot even collect 10 trajectories of data without falling to perform one policy gradient update.
And Recovery RL requires much more steps to learn since its safety critic, pre-trained with the objective function to maximize the cost values in simulation, is too conservative in predicting the cost violations in the real world.
%
%
For instance, in the catwalk task our approach achieves \textit{zero} falls and uses $\pi_{\rm safe}$ 6.8\% of total steps among first policy update, compared to Recovery RL with zero falls but 75.6\% of total steps of use of $\pi_{\rm safe}.$ 
%
%
%
%
This makes these baselines laborious to train in practice.
Overall, we see that on these tasks, the reward increases and the percentage of uses of $\pi_{\rm safe}$ decreases over policy updates.
For instance, the percentage of uses of $\pi_{\rm safe}$ decreases from 20\% to near 0\% in the efficient gait task.
For the two-leg balance task, the percentage drops from near 82.5\% to 67.5\%, suggesting that the two-leg balance is substantially harder than the previous two tasks. 
Still, the policy does improve the reward.
This observation implies that the learner can gradually learn the task while avoiding triggering $\pi_{\rm safe}.$
In addition, this suggests that we can design a $\mathcal{C}_{\rm tri}$ and $\pi_{\rm safe}$ that do not hinder the exploration of the policy as the performance increases.
%

%
%
%
%
%
%
%

%
Finally, Fig. \ref{fig:algorithm} shows the results of learned locomotion skills (please visit our project website for more videos). 
First, in the efficient gait task, the robot learns to use a smaller stepping frequency and achieves 34\% less energy than the nominal trotting gait.
Second, in the catwalk task, the distance between two sides of the legs is 0.09m, which is 40.9\% smaller than the nominal distance. 
Third, in the two-leg balance task, the robot can maintain balance by jumping up to four times via two legs, compared to two jumps from the policy pre-trained from simulation.
Without $\pi_{\rm safe},$ learning such locomotion skills would damage the robot and require manually re-positioning the robot when falling.
Note that there is \textit{no single fall} nor \textit{a manual reset} during the entire learning in the efficient gait (45mins of real-world training data collection, excluding automatic position reset or battery replacement) and catwalk tasks (29mins), and less than 5 falls in the two-leg balance task (28mins).
Our results suggest that learning legged locomotion skills autonomously is possible in the real world.

\section{Conclusions}
We studied the problem of safe reinforcement learning for acquiring locomotion skills for quadruped robots by combining a safe recovery policy and a learner policy.
%
%
The safe recovery policy takes over the control when the robot is close to a safety violation, and returns the control back when the robot stays safe in the estimated near future.
We provided a theoretical analysis of the algorithm and quantified the effect on the model errors for the learning performance. 
We evaluate our algorithm on both simulated and real quadruped robots for a variety of challenging locomotion tasks and demonstrate the effectiveness of the proposed framework compared to baseline methods.
%

%
No model is without limitation.
Although uncommon, our current $\pi_{\rm safe}$ can still fail to save the robot from unsafe states (see supplementary video for failure case) due to imperfections in $\pi_{\rm safe}$ and the safety trigger set $\mathcal{C}_{\rm tri}$. Further investigations in improving these would enable the learner policy to explore the environment more effectively and thus improve the learning efficiency.
In addition, we currently do not consider the model uncertainty from the environment and non-linear dynamics in our theoretical analysis. Including these would further improve the generality of our approach.
Finally, designing an appropriate reward when incorporating the $\pi_{\rm safe}$ can impact our learning performance. We use a penalty-based approach that obtained reasonable results in our experiments. We plan to investigate this in future work to further improve the learning performance.

\addtolength{\textheight}{-2cm}   



\section*{ACKNOWLEDGMENT}
The authors would like to thank Professor Peter J. Ramadge for the helpful discussion and members of Google Research for helpful discussion and support of the experiments.


\bibliographystyle{IEEEtran}
\bibliography{IEEEabrv_v4}

\end{document}